\documentclass[letterpaper]{article} 
\usepackage{aaai24}  
\usepackage{times}  
\usepackage{helvet}  
\usepackage{courier}  
\usepackage[hyphens]{url}  
\usepackage{graphicx} 
\usepackage{natbib}  
\usepackage{caption} 
\usepackage{algorithm}
\usepackage{algorithmic}

\usepackage{newfloat}
\usepackage{listings}
\DeclareCaptionStyle{ruled}{labelfont=normalfont,labelsep=colon,strut=off} 
\floatstyle{ruled}
\newfloat{listing}{tb}{lst}{}
\floatname{listing}{Listing}
\newcommand{\ie}{\textit{i}.\textit{e}.}
\newcommand{\eg}{\textit{e}.\textit{g}.}

\usepackage{subfig}
\usepackage{amsmath}
\usepackage{amsfonts}
\usepackage{dsfont}
\usepackage{enumitem}
\usepackage{multirow}
\usepackage{makecell}
\usepackage{subfloat}
\usepackage{xcolor}

\usepackage[colorlinks=true]{hyperref} 
\usepackage{navigator} 

\title{GMMFormer: Gaussian-Mixture-Model Based Transformer for Efficient \\ Partially Relevant Video Retrieval}
\author {
    Yuting Wang\textsuperscript{\rm 1,3},
    Jinpeng Wang\textsuperscript{\rm 1,3},
    Bin Chen\textsuperscript{\rm 2,3}\thanks{Corresponding author.},
    Ziyun Zeng\textsuperscript{\rm 1,3},
    Shu-Tao Xia\textsuperscript{\rm 1,3}
}
\affiliations {
    \textsuperscript{\rm 1} Tsinghua Shenzhen International Graduate School, Tsinghua University\\
    \textsuperscript{\rm 2} Harbin Institute of Technology, Shenzhen\\
    \textsuperscript{\rm 3} Research Center of Artificial Intelligence, Peng Cheng Laboratory\\
    \{wangyt22, wjp20, zengzy21\}@mails.tsinghua.edu.cn, chenbin2021@hit.edu.cn, xiast@sz.tsinghua.edu.cn
}

\usepackage{bibentry}

\begin{document}

\maketitle

\begin{abstract}
    Given a text query, partially relevant video retrieval (PRVR) seeks to find untrimmed videos containing pertinent moments in a database. For PRVR, clip modeling is essential to capture the partial relationship between texts and videos. Current PRVR methods adopt scanning-based clip construction to achieve explicit clip modeling, which is information-redundant and requires a large storage overhead. To solve the efficiency problem of PRVR methods, this paper proposes GMMFormer, a \textbf{G}aussian-\textbf{M}ixture-\textbf{M}odel based Trans\textbf{former} which models clip representations implicitly. During frame interactions, we incorporate Gaussian-Mixture-Model constraints to focus each frame on its adjacent frames instead of the whole video. Then generated representations will contain multi-scale clip information, achieving implicit clip modeling. In addition, PRVR methods ignore semantic differences between text queries relevant to the same video, leading to a sparse embedding space. We propose a query diverse loss to distinguish these text queries, making the embedding space more intensive and contain more semantic information. Extensive experiments on three large-scale video datasets (\ie, TVR, ActivityNet Captions, and Charades-STA) demonstrate the superiority and efficiency of GMMFormer. Code is available at \url{https://github.com/huangmozhi9527/GMMFormer}.
\end{abstract}

\section{Introduction}

\begin{figure}[t]
  \centering
  \includegraphics[width=\linewidth]{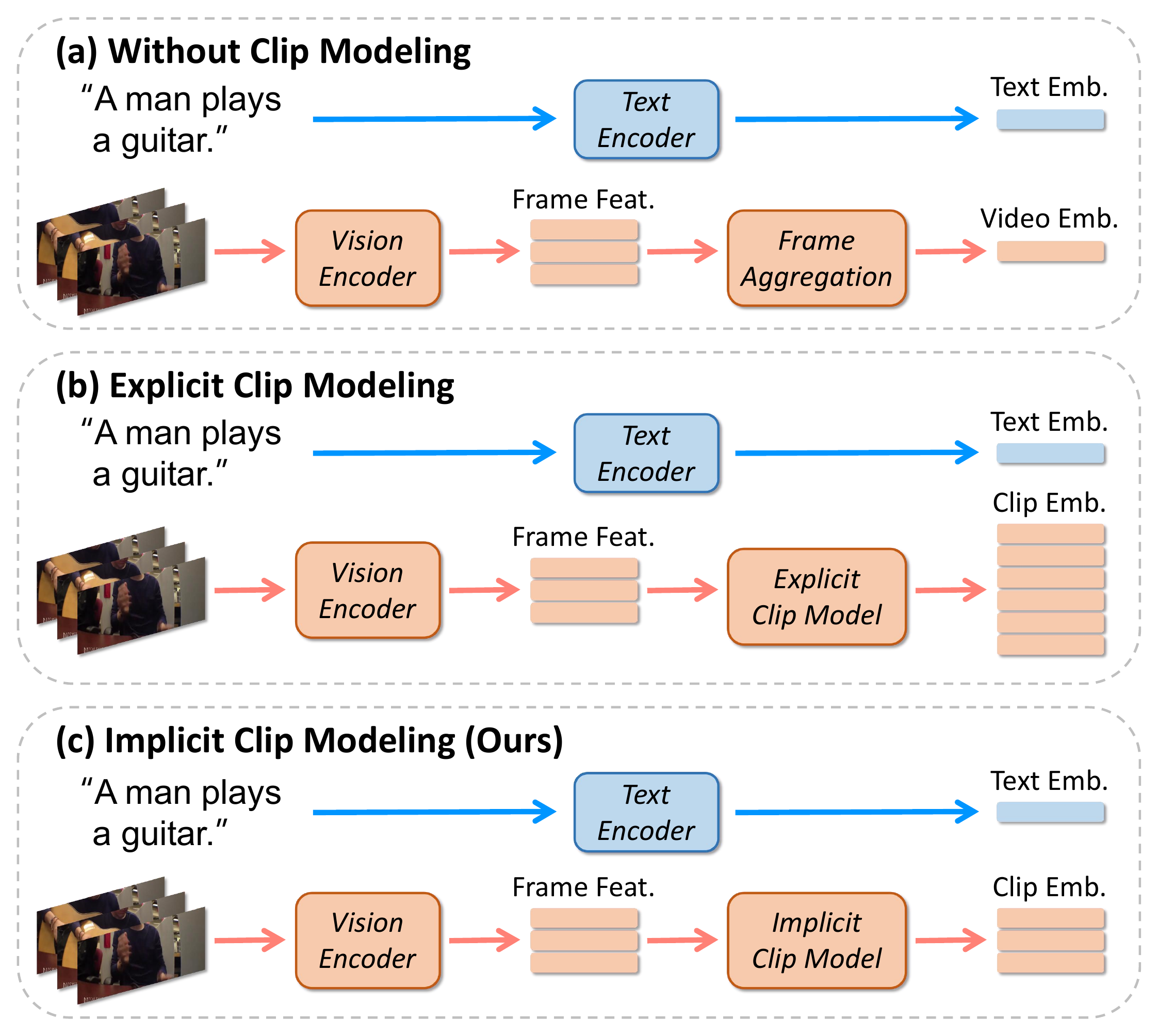}
  \caption{Traditional text-to-video retrieval pipelines (a) generate compact video embeddings and lost clip information. Previous partially relevant video retrieval pipelines (b) adopt explicit clip modeling, which is information-redundant and requires a large storage overhead. We utilize implicit clip modeling (c) to obtain compact clip embeddings, containing multi-scale clip information.}
  \label{intro}
\end{figure}

With the development of society, video has become the subject of information dissemination. As a result, text-to-video retrieval (T2VR) \cite{dong2018predicting, chen2020fine, miech2019howto100m, liu2019use, li2019w2vv++, faghri2017vse++, dong2019dual, dong2021dual, dong2022reading} has received increasing attention from academia and industry. Given a text query, T2VR aims to retrieve semantically relevant videos from a video database. However, videos in T2VR datasets are pre-trimmed to be entirely relevant to corresponding text queries, which exists a gap from the real world. In realistic social media or video platforms (\eg, YouTube), a video is usually long-time and contains several moments, among which only one moment is entirely relevant to the corresponding text query. When handling these untrimmed videos, T2VR models trained on pre-trimmed video datasets may not perform well, resulting in poor user experience. To overcome the above-mentioned problem, ~\cite{dong2022partially} proposed the partially relevant video retrieval (PRVR) task, which collects untrimmed videos to form the video database. In particular, a video in PRVR corresponds to several text queries, and a text query is only relevant to one moment within the video. Compared to T2VR, PRVR is more aligned with the natural world and has more research significance.

Given a text-video pair, previous PRVR methods employ pre-trained vision-language models to extract frame and word features. These features will pass through sequential models (\eg, RNN, LSTM, Transformer \cite{vaswani2017attention}, etc.) to model global sequential interactions, generating frame and sentence embeddings. After that, they model clip representations to capture the partial relationship between the text and video. Specifically, a multi-scale sliding window strategy is applied to frame embeddings to construct clip embeddings. Finally, the text-video similarity can derive from similarities between sentence embeddings with clip and frame embeddings.

Those PRVR methods have outperformed T2VR methods on untrimmed video datasets. However, their retrieval pipelines still suffer from two problems. \textbf{1) Global frame interactions confuse different moments of untrimmed videos.} An untrimmed video contains several moments. These moments correspond to different text queries, which the PRVR model should distinguish. However, we find that global frame interactions will make frame embeddings similar to each other. With these similar embeddings, the model cannot locate the correct time period of the given text query, resulting in poor performance. \textbf{2) Explicit clip modeling by scanning-based clip construction is information-redundant and requires a large storage overhead.} The multi-scale sliding window strategy will traverse all possible clips, generating a lot of irrelevant clip embeddings and leading to information redundancy. With frame embeddings of length $M$, the generated clip embeddings will have a length of $M(M+1) / 2$. For instance, the past SOTA PRVR method MS-SL \cite{dong2022partially} downsamples frame features to 32-length and constructs 528-length clip embeddings, within which only five clips are relevant to corresponding text descriptions on the TVR dataset. Although these redundant clip embeddings make the model localize the time period more accurately, they require a large storage overhead and reduce retrieval efficiency. 

To solve the above-mentioned two problems, in this paper, we propose GMMFormer, a Gaussian-Mixture-Model based Transformer to model clip representations implicitly. Our motivation lies in a natural characteristic: moments in a video are successive and have limited duration, within which each frame should pay more attention to its neighboring frames; the closer it is, the more attention should be paid. Inspired by \cite{fu2022geo, qu2020context, zhou2023dual, kim2020t}, we design a GMMFormer block to incorporate Gaussian-Mixture-Model constraints during frame interactions to focus each frame on its adjacent frames. In particular, we utilize multi-scale Gaussian windows to model frame interactions of different ranges, generating clip features with several receptive fields. Then we aggregate these features to obtain clip embeddings. These clip embeddings contain multi-scale clip information and can perceive video moments with different lengths. The comparison of different retrieval pipelines is illustrated in Figure \ref{intro}.

For a video in PRVR, its relevant text queries are semantically diverse. However, the commonly used triplet ranking loss \cite{dong2021dual, faghri2017vse++} and infoNCE loss \cite{miech2020end, zhang2021video} treat them equally and pull them together in the embedding space. These losses disturb the semantic structure of text representations, thus resulting in a sparse distribution in the embedding space. In this paper, we propose a query diverse loss to distinguish text queries relevant to the same video. Inspired by \cite{wang2020understanding}, given an untrimmed video, we push its relevant text queries away from each other, generating discriminative sentence embeddings. Then the embedding space will be more intensive and contain more semantic information.

We conducted extensive experiments on three large-scale video datasets TVR \cite{lei2020tvr}, ActivityNet Captions \cite{krishna2017dense}, and Charades-STA \cite{gao2017tall}. The experimental results demonstrate the superiority and efficiency of our GMMFormer. In particular, GMMFormer achieves state-of-the-art results on three datasets. And compared to the past SOTA MS-SL, GMMFormer is about 2.5 times faster and the storage overhead is 20 times smaller.

Overall, our main contributions are as follows:

\begin{itemize}[leftmargin=*]
    \item We propose GMMFormer, a Gaussian-Mixture-Model based Transformer to model clip representations implicitly. GMMFormer is effective for its multi-scale Gaussian constraints and efficient for its compact clip embeddings with high information density.
    \item We propose a query diverse loss to distinguish different text queries relevant to the same video, preserving the semantic structure of text representations.
    \item Extensive experiments and ablation studies on three large-scale datasets (\ie, TVR, ActivityNet Captions, Charades-STA) demonstrate the superiority and efficiency of our GMMFormer.
\end{itemize}

\section{Related Work}

\textbf{Text-to-video Retrieval.} Video analysis \cite{wang2023contrastive, wang2022hugs, zeng2022motion, liu2023cav3, liu2023mobile, jin2022you} has recently gained much attention due to the increasing video data on the Internet. Among them, the text-to-video retrieval (T2VR) task \cite{dong2018predicting, chen2020fine, li2019w2vv++, faghri2017vse++, gao2023ciba, lei2021detecting, li2023momentdiff} aims to retrieve relevant videos from a set of pre-trimmed video clips given a text description. A standard pipeline for T2VR is to first encode videos and texts to obtain video and sentence representations, and then map them into a common embedding space to measure the cross-modal similarity. 

\textbf{Partially Relevant Video Retrieval.} The partially relevant video retrieval (PRVR) task \cite{dong2022partially} aims to retrieve untrimmed videos partially relevant to a given query, which is more in line with the real world than T2VR. For PRVR, clip modeling is crucial in capturing the partial relationship between texts and videos. Previous PRVR methods adopt clip construction to achieve explicit clip modeling. They apply a multi-scale sliding window strategy on frame embeddings to obtain clip embeddings. This practice will traverse all possible clips and generate a lot of irrelevant clip embeddings, requiring a large storage overhead and reducing retrieval efficiency. Besides, PRVR models are easy to overfit, which might be improved by adversarial training \cite{gao2023ciba, bai2021improving, bai2020improving, gudibandetest}. In this paper, we propose GMMFormer, a Gaussian-Mixture-Model based Transformer to model clip representations implicitly. GMMFormer can generate compact clip embeddings with high information density, which is effective and efficient.

\textbf{Video Corpus Moment Retrieval.} The video corpus moment retrieval (VCMR) task \cite{song2021spatial, lei2020tvr} seeks to retrieve moments semantically relevant to a given query from a collection of untrimmed videos. VCMR methods adopt a two-stage pipeline. They retrieve several candidate videos in the first stage, which may contain the target moment, then retrieve moments from the candidate videos in the second stage. VCMR’s first stage is similar to PRVR. However, VCMR requires moment-level annotations, which is time-consuming and labor-intensive.

\section{Methodology}

We explain in detail our approach for PRVR. We start with the formulation of PRVR in Section \ref{sec31}, then elaborate on the overview of GMMFormer in Section \ref{sec34}. Next, we introduce our designed GMMFormer block in Section \ref{sec33} and the learning strategy in Section \ref{sec35}. 

\subsection{Problem Formulation}\label{sec31}

Given a text query, partially relevant video retrieval (PRVR) aims to retrieve videos containing a moment semantically relevant to the given query, from a large corpus of untrimmed videos. Each video in PRVR databases has several moments and is associated with multiple text descriptions, while each text description represents the content of a specific moment in the corresponding video. It is worth mentioning that the start or end time points of moments are unavailable in PRVR.

\subsection{Overview}\label{sec34}

\begin{figure}[t]
  \centering
  \includegraphics[width=\linewidth]{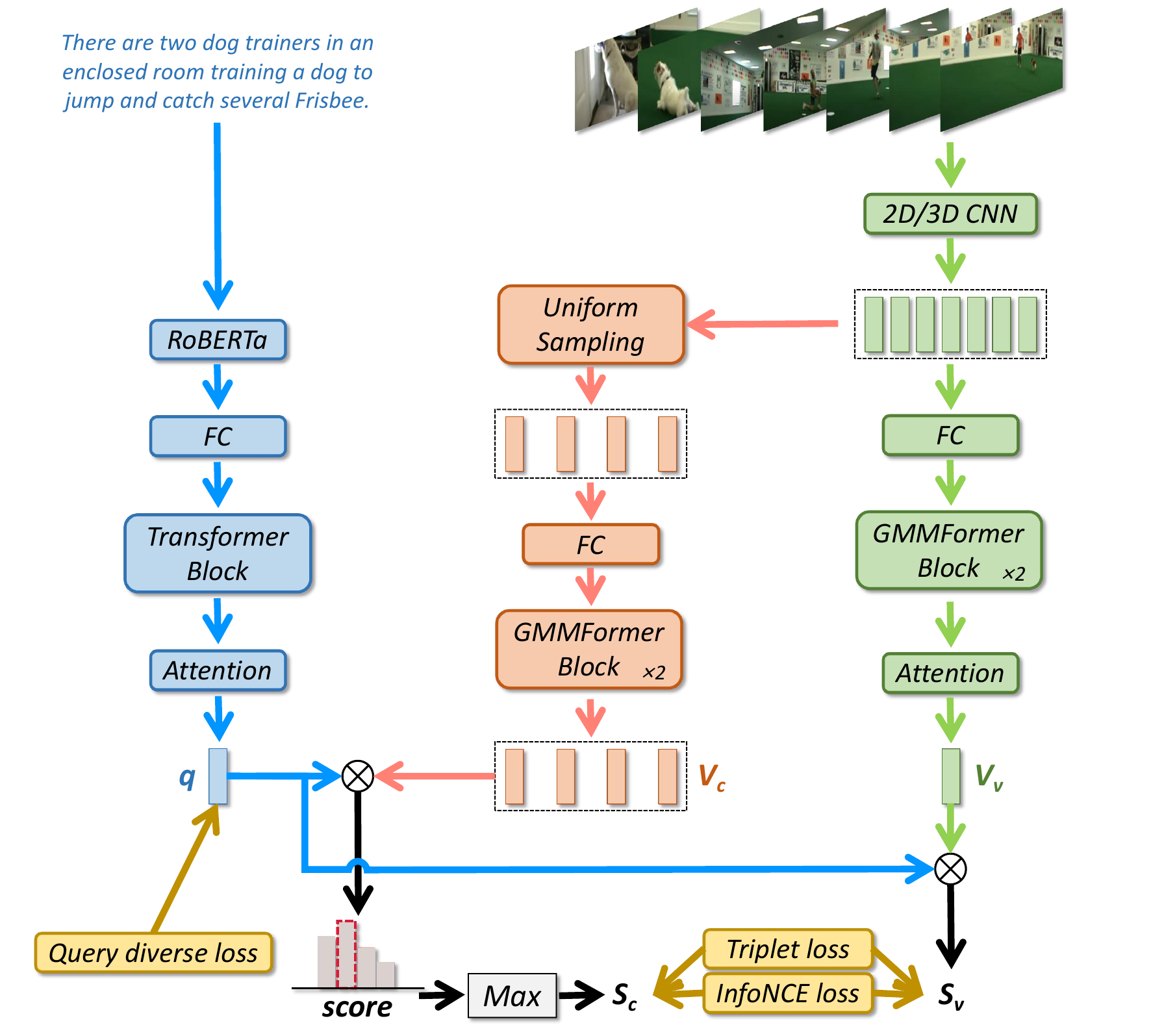}
  \caption{\textbf{The overall framework of GMMFormer.} $\otimes$ denotes the matrix multiplication.}
  \label{overview}
\end{figure}

In this section, we introduce the overall framework of our GMMFormer, including sentence representation encoding, video representation encoding and similarity measure, as shown in Figure \ref{overview}.

\textbf{Sentence Representation.} Given a sentence containing $N$ words, we first utilize a pre-trained RoBERTa \cite{liu2019roberta} to extract word features. Then we adopt a FC layer with a ReLU activation to embed the word features into a lower-dimensional space. After adding the learnable positional embedding to the mapped features, we employ a vanilla Transformer layer to obtain a sequence of $d$-dimensional contextualized word feature vectors $Q = \{q_i\}_{i=1}^{N} \in \mathbb{R}^{N\times d}$. It is worth mentioning that we do not use the GMMFormer block here, which is designed for untrimmed videos. Finally, we use a simple attention module on $Q$ to obtain sentence embeddings $q \in \mathbb{R}^d$:
\begin{gather}
q = \sum_{i=1}^{N} a_i^q \times q_i, a^q = softmax(w Q^T)
\end{gather}
where $w \in \mathbb{R}^{1\times d}$ is a trainable vector and  $a^q \in \mathbb{R}^{1\times N}$ indicates the attention vector.

\textbf{Video Representation.} Given an untrimmed video containing $M_f$ frames, we first employ a pre-trained 2D or 3D CNN to extract frame features. Then we pass them through two branches to obtain clip and video embeddings. Clip embeddings help model to locate relevant moments, while video embeddings measure the global text-video similarity.

In the clip-level branch, we uniformly sample a fixed number of feature vectors by mean pooling over the corresponding multiple consecutive frame features. Then we use a FC layer with a ReLU activation to reduce dimension, obtaining clip features. Finally, we use two GMMFormer blocks with the learnable positional embedding on clip features to get clip embeddings $V_c = \{c_i\}_{i=1}^{Mc} \in \mathbb{R}^{M_c\times d}$, where $M_c$ is the sampled number and $d$ is the dimension.

In the video-level branch, similarly, we first use a FC layer with a ReLU activation to reduce dimension, then employ two GMMFormer layers with the learnable positional embedding to obtain contextualized features $V_f = \{v_i\}_{i=1}^{M_f} \in \mathbb{R}^{M_f\times d}$. Finally, we employ a simple attention module on $V_f$ to obtain video embeddings $V_v \in \mathbb{R}^d$:
\begin{gather}
V_v = \sum_{i=1}^{M_f} a_i^f \times v_{i}, a^f = softmax(w V_f^T)
\end{gather}
where $w \in \mathbb{R}^{1\times d}$ is a trainable vector and  $a^f \in \mathbb{R}^{1\times M_f}$ indicates the attention vector.

\textbf{Similarity Measure.} Given a text-video pair, we first compute the above-mentioned $q, V_c, V_v$, then the video-level similarity is measured as the cosine similarity between sentence embeddings $q$ and video embeddings $V_v$:
\begin{gather}
S_v(t, v) = cos(q, V_v)
\end{gather}

Besides, we use the cosine similarity and max-pooling operation to calculate the clip-level similarity between sentence embeddings $q$ and clip embeddings $V_c$:
\begin{gather}
S_c(t, v) = max \{cos(q, c_1), ..., cos(q, c_{M_c})\}
\end{gather}

The similarity of the text-video pair can be computed as the weighted sum of the video-level similarity and clip-level similarity:
\begin{gather}
S(t, v) = \alpha_v S_v(t, v) + \alpha_c S_c(t, v)
\end{gather}
where $\alpha_v, \alpha_c \in [0, 1]$ are hyper-parameters to balance two similarities, and $\alpha_v + \alpha_c = 1$.

\subsection{GMMFormer Block}\label{sec33}

\begin{figure}[t]
  \centering
  \includegraphics[width=\linewidth]{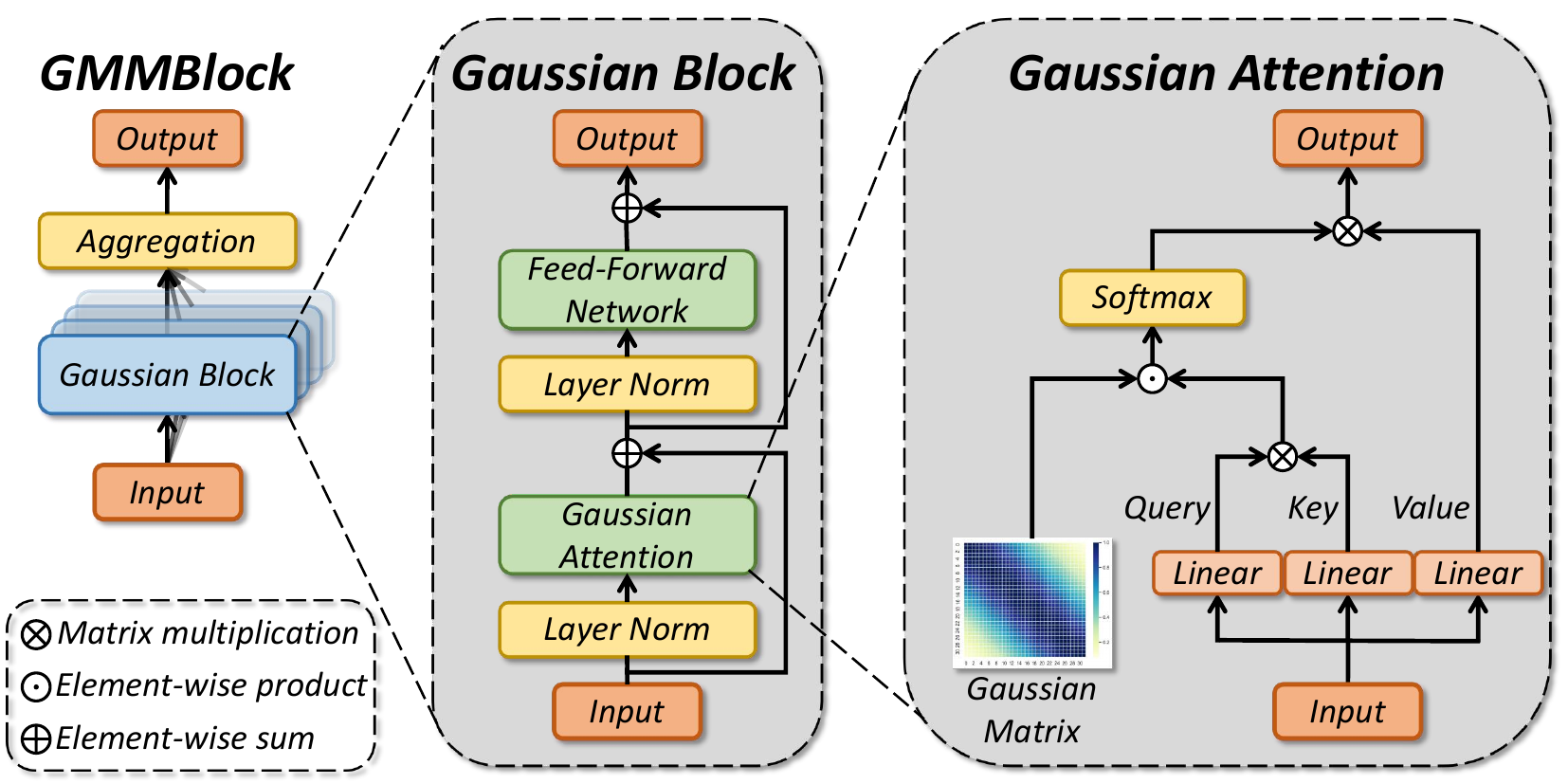}
  \caption{\textbf{The illustration of the GMMFormer block.}}
  \label{block}
\end{figure}

To model the Gaussian-Mixture-Model distribution of video representations, we first propose a Gaussian block to incorporate a Gaussian constraint during frame interactions. Then we employ multi-scale Gaussian blocks in parallel and aggregate their output, making it a Gaussian-Mixture-Model constraint, as shown in Figure \ref{block}.

Given $M$ extracted features, we present it in a matrix form $X_i \in \mathbb{R}^{M\times d}$, where $d$ is the feature dimension and $i$ is the video index. In our designed Gaussian block, we project the input matrix $X_i$ to three matrices query, key and value via three learnable parameters $W^q$, $W^k$ and $W^v$. We use the query matrix to perform scaled dot-product attention over the key matrix, obtaining an attention score matrix. Then we design a Gaussian matrix $W^g \in \mathbb{R}^{M\times M}$ composed of $M$ Gaussian windows to perform element-wise product over the attention score matrix. After that, we put the generation through a softmax function to determine attentional distributions over the value matrix. The resulting weight-averaged value matrix forms the output of the Gaussian attention module in the Gaussian block:
\begin{gather}
    X_i^{attn} = softmax(W^g \odot \frac{X_i W^q (X_i W^k)^T}{\sqrt{d_k}})X_i W^v \\
    W^g(i, j) = \frac{1}{2\pi} e^{-\frac{(j-i)^2}{\sigma^2}}
\end{gather}
where $d_k$ is the dimension of queries and keys,  $\sigma^2$ is the variance of the Gaussian density distribution and $\odot$ indicates the element-wise product function. 

After the Gaussian attention module, we feed $X_i^{attn}$ to a Feed-Forward Network (FFN) to obtain Gaussian block output $X_i^{output}$. Similar to the vanilla Transformer block, we add residual connection \cite{he2016deep} and Layer Normalization \cite{ba2016layer} in the Gaussian attention module and the FFN module. So the Gaussian block can be formulated as:
\begin{gather}
    X_i^{output} = FFN(LayerNorm(X_i^{inter})) + X_i^{inter} \\
    X_i^{inter} = GauAttn(LayerNorm(X_i)) + X_i
\end{gather}
where $GauAttn$ indicates the Gaussian attention module, and FFN is composed of two fully connected (FC) layers.

Gaussian block output will contain fixed-length clip information. However, video moments are diverse in length. So we employ multi-scale Gaussian blocks in parallel and aggregate their output. Here, we use average pooling to achieve aggregation:
\begin{gather}
X_i^{GMM} = \frac{1}{K} \sum_{k=1}^{K} GB(X_i, \sigma_{k}^{2})
\end{gather}
where $GB(X_i, \sigma_{k}^{2})$ is a Gaussian block with the variance $\sigma_{k}^{2}$ and $K$ is the number of Gaussian blocks. Specifically, we set $K=4$ and choose Gaussian blocks respectively with low, medium, high, and infinite variance. $X_i^{GMM}$ denotes the output of the GMMFormer block, which maintains the length of $M$ and contains multi-scale clip information. 

\subsection{Learning}\label{sec35}

We consider a text-video pair positive if the video contains a moment relevant to the text and negative if there is no relevant content. We adopt triplet ranking loss \cite{dong2021dual, faghri2017vse++} and infoNCE loss \cite{miech2020end, zhang2021video} that are widely used in the retrieval task. 

Given a positive text-video pair $(t, v)$, the triplet ranking loss over the mini-batch $\mathcal{B}$ is defined as:
\begin{gather}
\mathcal{L}^{trip} = \frac{1}{n} \sum_{(t,v) \in \mathcal{B}} \{max(0, m + S(t^-, v) - S(t, v)) \nonumber \\ 
    + max(0, m + S(t, v^-) - S(t,v))\}
\end{gather}
where $m$ is a margin constant, $t^-$ and $v^-$ indicate a negative text for $v$ and a negative video for $t$. Similar to \cite{dong2022partially}, we randomly sample the negative samples from the mini-batch at the beginning of the training and choose the hardest negative samples after 20 epochs.

Given a positive text-video pair $(t, v)$, the infoNCE loss over the mini-batch $\mathcal{B}$ is computed as:
\begin{gather}
\mathcal{L}^{nce} = -\frac{1}{n} \sum_{(t,v) \in \mathcal{B}} \{log(\frac{S(t,v)}{S(t,v) + \sum\nolimits_{t_i^{-} \in \mathcal{N}_t }S(t_i^-, v)}) \nonumber\\
+ log(\frac{S(t,v)}{S(t,v) + \sum\nolimits_{v_i^{-} \in \mathcal{N}_v }S(t, v_i^-)}) \}
\end{gather}
where $\mathcal{N}_t$ denotes all negative texts of the video $v$ in the mini-batch, while $\mathcal{N}_v$ denotes all negative videos of the query $t$ in the mini-batch.

Besides, given a collection of texts $T$ in a mini-batch, we design a query diverse loss to distinguish different text queries relevant to the same video, defined as:
\begin{gather}
\mathcal{L}^{div} = \frac{1}{n} \sum_{t_i,t_j \in T} \mathds{1}_{t_i, t_j} log(1 + e^{\alpha (cos(t_i, t_j) + \delta)})
\end{gather}
where $\delta > 0$ is a margin, $\alpha > 0$ is a scaling factor and $\mathds{1}_{t_i, t_j} \in \{0,1\}$ is an indicator function. $\mathds{1}_{t_i, t_j} = 1$ when $t_i$ and $t_j$ are relevant to the same video.

$\mathcal{L}^{div}$ will push away semantically diverse texts relevant to the same video, preserving the semantic structure of text representations. Then the embedding space will be more intensive and contain more semantic information.

Finally, our model is trained by minimizing the following overall training loss:
\begin{gather}
\mathcal{L} = \mathcal{L}_{c}^{trip} + \mathcal{L}_{v}^{trip} + \lambda_1 \mathcal{L}_{c}^{nce} + \lambda_2 \mathcal{L}_{v}^{nce} + \lambda_3 \mathcal{L}^{div}
\end{gather}
where $\mathcal{L}_{c}^{trip}$ and $\mathcal{L}_{v}^{trip}$ denote the triplet ranking losses using the clip-level similarity $S_c$ and video-level similarity $S_v$, and accordingly for $\mathcal{L}_{c}^{nce}$ and $\mathcal{L}_{v}^{nce}$. $\lambda_1$, $\lambda_2$ and $\lambda_3$ are hyper-parameters to balance corresponding losses.

\begin{table} [tb!]
\caption{Performance of various models on the TVR dataset. 
Models are sorted in ascending order in terms of their SumR.
}
\label{tab:sota-tvr}
\centering 
\scalebox{0.92}{
\begin{tabular}{l|c|c|c|c|c}
\hline
\textbf{Model} & \textbf{R@1} & \textbf{R@5} & \textbf{R@10} & \textbf{R@100} & \textbf{SumR}\\
\hline
\multicolumn{6}{l}{\emph{T2VR models:}}  \\
\hline
W2VV & 2.6 & 5.6 & 7.5 & 20.6 & 36.3\\
HGR & 1.7 & 4.9 & 8.3 & 35.2 & 50.1  \\
HTM  & 3.8 & 12.0 & 19.1 & 63.2 & 98.2\\
CE & 3.7 & 12.8 & 20.1 & 64.5 & 101.1  \\
DE++ & 8.8 & 21.9 & 30.2 & 67.4  & 128.3 \\
RIVRL & 9.4 & 23.4 & 32.2 & 70.6 & 135.6\\
CLIP4Clip & 9.9 & 24.3 & 34.3 & 72.5 & 141.0\\
Cap4Video & 10.3 & 26.4 & 36.8 & 74.0 & 147.5\\
\hline
\multicolumn{6}{l}{\emph{VCMR models w/o moment localization:}}  \\
\hline
XML & 10.0 & 26.5 & 37.3 & 81.3  & 155.1\\
ReLoCLNet & 10.7 & 28.1 & 38.1 & 80.3  & 157.1 \\ 
CONQUER & 11.0 & 28.9 & 39.6 & 81.3  & 160.8 \\ 
\hline
\multicolumn{6}{l}{\emph{PRVR models:}}  \\
\hline
MS-SL & 13.5 & 32.1 & 43.4 & 83.4 & 172.4\\
GMMFormer & \textbf{13.9} & \textbf{33.3} & \textbf{44.5} & \textbf{84.9} & \textbf{176.6}\\
\hline
\hline
\end{tabular}
 }
\end{table}

\begin{table} [tb!]
\caption{Performance of various models on the ActivityNet Captions dataset.
}
\label{tab:sota-activitynet}
\centering 
\scalebox{0.92}{
\begin{tabular}{l|c|c|c|c|c}
\hline
\textbf{Model} & \textbf{R@1} & \textbf{R@5} & \textbf{R@10} & \textbf{R@100} & \textbf{SumR}\\
\hline
\multicolumn{6}{l}{\emph{T2VR models:}}  \\
\hline
W2VV & 2.2 & 9.5 & 16.6 & 45.5 & 73.8\\
HTM  & 3.7 & 13.7 & 22.3 & 66.2 & 105.9\\
HGR & 4.0 & 15.0 & 24.8 & 63.2 & 107.0  \\
RIVRL & 5.2 & 18.0 & 28.2 & 66.4 & 117.8\\
DE++ & 5.3 & 18.4 & 29.2 & 68.0 & 121.0 \\
CE & 5.5 & 19.1 & 29.9 & 71.1 & 125.6  \\
CLIP4Clip & 5.9 & 19.3 & 30.4 & 71.6 & 127.3\\
Cap4Video & 6.3 & 20.4 & 30.9 & 72.6 & 130.2\\
\hline
\multicolumn{6}{l}{\emph{VCMR models w/o moment localization:}}  \\
\hline
ReLoCLNet & 5.7 & 18.9 & 30.0 & 72.0 & 126.6 \\ 
XML & 5.3 & 19.4 & 30.6 & 73.1 & 128.4\\
CONQUER & 6.5 & 20.4 & 31.8 & 74.3  & 133.1 \\ 
\hline
\multicolumn{6}{l}{\emph{PRVR models:}}  \\
\hline
MS-SL & 7.1 & 22.5 & 34.7 & 75.8 & 140.1\\
GMMFormer & \textbf{8.3} & \textbf{24.9} & \textbf{36.7} & \textbf{76.1} & \textbf{146.0}\\
\hline
\hline
\end{tabular}
 }
\end{table}

\begin{table} [tb!]
\caption{Performance of various models on the Charades-STA dataset.
}
\label{tab:sota-charades}
\centering 
\scalebox{0.92}{
\begin{tabular}{l|c|c|c|c|c}
\hline
\textbf{Model} & \textbf{R@1} & \textbf{R@5} & \textbf{R@10} & \textbf{R@100} & \textbf{SumR}\\
\hline
\multicolumn{6}{l}{\emph{T2VR models:}}  \\
\hline
W2VV & 0.5 & 2.9 &4.7 & 24.5 & 32.6\\
HGR & 1.2 & 3.8 & 7.3 & 33.4 & 45.7  \\
CE & 1.3 & 4.5 & 7.3 & 36.0 & 49.1  \\
DE++& 1.7 & 5.6 & 9.6 & 37.1  & 54.1 \\
RIVRL  & 1.6 & 5.6 & 9.4 & 37.7 & 54.3\\ 
HTM  & 1.2 & 5.4 & 9.2 & 44.2 & 60.0\\
CLIP4Clip & 1.8 & 6.5 & 10.9 & 44.2 & 63.4\\
Cap4Video & 1.9 & 6.7 & 11.3 & 45.0 & 65.0\\
\hline
\multicolumn{6}{l}{\emph{VCMR models w/o moment localization:}}  \\
\hline
ReLoCLNet & 1.2 & 5.4 & 10.0 & 45.6 & 62.3\\
XML & 1.6 & 6.0 & 10.1 & 46.9 & 64.6\\
CONQUER &1.8 & 6.3 & 10.3 & 47.5  & 66.0 \\ 
\hline
\multicolumn{6}{l}{\emph{PRVR models:}}  \\
\hline
MS-SL & 1.8 & 7.1 & 11.8 & 47.7 & 68.4\\
GMMFormer & \textbf{2.1} & \textbf{7.8} & \textbf{12.5} & \textbf{50.6} & \textbf{72.9}\\
\hline
\hline
\end{tabular}
 }
\end{table}

\begin{table*} [tb!]
\caption{Model comparisons in terms of FLOPs and parameters.
}
\label{tab:flops}
\centering 
\scalebox{0.92}{
\begin{tabular}{l|c|c|c|c|c}
\hline
 & \textbf{CLIP4Clip} & \textbf{Cap4Video} & \textbf{CONQUER} & \textbf{MS-SL}& \textbf{GMMFormer}\\
\hline
\textbf{FLOPs (G)} & 5.77 & 7.35 & 5.65 & 1.29 & 1.95\\
\textbf{Params (M)} & 103.65 & 104.84 & 22.55 & 4.85 & 12.85\\
\hline
\hline
\end{tabular}
 }
\end{table*}

\begin{table} [tb!]
\caption{Comparisons in terms of retrieval efficiency of PRVR models.
}
\label{tab:efficiency}
\centering 
\scalebox{0.88}{
\begin{tabular}{l|c|c|c|c|c}
\hline
\textbf{Database Size} & \textbf{500} & \textbf{1,000} & \textbf{1,500} & \textbf{2,000} & \textbf{2,500}\\
\hline
\multicolumn{6}{l}{\emph{runtime (ms):}}  \\
\hline
MS-SL & 4.89 & 6.11 & 8.06 & 10.42 & 12.93\\
GMMFormer & \textbf{2.68} & \textbf{2.93} & \textbf{3.40} & \textbf{3.94} & \textbf{4.56}\\
\hline
\multicolumn{6}{l}{\emph{memory usage (M):}}  \\
\hline
MS-SL & 50.02 & 100.04 & 150.06 & 200.08 & 250.11\\
GMMFormer & \textbf{2.53} & \textbf{5.07} & \textbf{7.60} & \textbf{10.14} & \textbf{12.67}\\
\hline
\hline
\end{tabular}
 }
\end{table}

\section{Experiments}

\subsection{Experimental Setup}

\textbf{Datasets.} We evaluate our GMMFormer on three large-scale video datasets (\ie, TV show Retrieval (TVR) \cite{lei2020tvr}, ActivityNet Captions \cite{krishna2017dense}, and Charades-STA \cite{gao2017tall}). Note that moment annotations provided by these datasets are unavailable in the PRVR task. \textbf{TVR} contains 21.8K videos collected from 6 TV shows. Five natural language sentences are associated with each video, describing different moments in the video. Following \cite{dong2022partially}, we utilize 17,435 videos with 87,175 moments for training and 2,179 videos with 10,895 moments for testing. \textbf{ActivityNet Captions} has around 20K videos from YouTube. On average, each video has about 3.7 moments with corresponding sentence descriptions. We use the popular data partition used in \cite{zhang2021video, zhang2020hierarchical}. \textbf{Charades-STA} includes 6,670 videos with 16,128 sentence descriptions. Each video holds around 2.4 moments with corresponding text queries on average. We use the official data partition for model training and testing.

\textbf{Baselines.} Except the SOTA PRVR model MS-SL~\cite{dong2022partially}, we also compare our GMMFormer with models designed for T2VR and VCMR. In particular, we choose the following eight T2VR models, \ie, W2VV~\cite{dong2018predicting}, CE~\cite{liu2019use}, HTM \cite{miech2019howto100m}, HGR~\cite{chen2020fine}, DE++~\cite{dong2021dual}, RIVRL~\cite{dong2022reading}, CLIP4Clip~\cite{luo2022clip4clip}, Cap4Video~\cite{wu2023cap4video}, and the following three VCMR models, \ie, XML~\cite{lei2020tvr}, ReLoCLNet~\cite{zhang2021video}, CONQUER~\cite{hou2021conquer}.
These VCMR models are two-stage, where a first-stage module retrieves candidate videos, followed by a second-stage module to localize specific moments in the candidate videos. As moment annotations are unavailable in PRVR, we have re-trained VCMR models (removing their moment localization modules) using the same video features as ours. For Cap4Video, we utilize the manual crawling approach to obtain auxiliary captions.

\textbf{Evaluation Protocols.} Following \cite{dong2022partially}, we utilize rank-based metrics, namely $R$@$K$ ($K$ = 1, 5, 10, 100). $R$@$K$ is the fraction of queries that correctly retrieve desired items in the top $K$ of the ranking list. For overall comparisons, we also report the Sum of all Recalls (SumR).

\textbf{Implementation Details.} For video representations on TVR, we utilize features provided by \cite{lei2020tvr}, 3,072-D visual features obtained by concatenating frame-level ResNet152 \cite{he2016deep} features and segment-level I3D \cite{carreira2017quo} features. On ActivityNet Captions and Charades-STA, we only utilize I3D features provided by \cite{zhang2020hierarchical} and \cite{mun2020local}, respectively. For sentence representations, we use 768-D RoBERTa features provided by \cite{lei2020tvr} on TVR. On ActivityNet Captions and  Charades-STA, we use 1,024-D RoBERTa features extracted by \cite{dong2022partially}. For four types of Gaussian blocks (\ie, low, medium, high and infinite), we set the Gaussian variance to 0.5, 1.0, 5.0 and $\infty$ respectively.

\subsection{Main Results}

\textbf{Retrieval Performance.} Table \ref{tab:sota-tvr}, \ref{tab:sota-activitynet}, \ref{tab:sota-charades} report the retrieval performance of various models on three large-scale video datasets. As can be seen, T2VR models perform poorly compared to VCMR and PRVR models. They focus on the entire relevance between videos and texts, which makes great sense in the T2VR task but is sub-optimal for PRVR. VCMR models focus on retrieving moments, which to some extent, learn the partial relevance between videos and texts, leading to better performance than T2VR models. PRVR models have excellent performance, which is attributed to clip modeling. Among them, our GMMFormer achieves state-of-the-art performance. Major advantages in GMMFormer lie in 1) multi-scale Gaussian blocks enhance the ability to perceive different video moments, 2) and the query diverse loss preserves the semantic structure of text representations.

\textbf{Retrieval Efficiency.} In addition, we compare some competitive models mentioned above in terms of FLOPs and model parameters. As shown in Table \ref{tab:flops}, PRVR models are more lightweight than T2VR and VCMR models while achieving higher retrieval performance. Our GMMFormer has more parameters and calculations than MS-SL because of parallel Gaussian blocks. However, these Gaussian blocks are located in video branches, which will be offline to compute beforehand. We further compare GMMFormer with MS-SL regarding retrieval efficiency in an actual situation. Specifically, we build a video subset from TVR and measure average runtime and memory usage to complete the retrieval process for a single text query under different database sizes settings. For a fair comparison, the reported runtime is measured on the same Nvidia RTX3080Ti GPU. As shown in Table \ref{tab:efficiency}, GMMFormer is about 2.5 times faster than MS-SL, and the storage overhead of GMMFormer is 20 times smaller than MS-SL. The main superiority of GMMFormer in terms of efficiency lies in compact clip embeddings, which are generated by implicit clip modeling.

\begin{table} [tb!]
\caption{Ablation studies of GMMFormer on TVR. GB means GMMFormer block and QDL means query diverse loss.
}
\label{tab:ablation}
\centering 
\scalebox{0.92}{
\begin{tabular}{c|c|c|c|c|c|c}
\hline
\textbf{GB} & \textbf{QDL} & \textbf{R@1} & \textbf{R@5} & \textbf{R@10} & \textbf{R@100} & \textbf{SumR}\\
\hline
& & 11.6 & 29.6 & 40.4 & 81.8 & 163.5 \\
\checkmark & & 12.9 & 32.2 & 43.9 & 83.9 & 172.9 \\
& \checkmark & 12.3 & 31.4 & 42.5 & 83.6 & 169.9 \\
\hline
\checkmark & \checkmark & \textbf{13.9} & \textbf{33.3} & \textbf{44.5} & \textbf{84.9} & \textbf{176.6}\\
\hline
\hline
\end{tabular}
 }
\end{table}

\begin{table} [tb!]
\caption{Ablation studies of the constraint window on TVR. CW means constraint window.
}
\label{tab:window}
\centering 
\scalebox{0.92}{
\begin{tabular}{l|c|c|c|c|c}
\hline
\textbf{CW} & \textbf{R@1} & \textbf{R@5} & \textbf{R@10} & \textbf{R@100} & \textbf{SumR}\\
\hline
Boxcar & 12.9 & 32.1 & 43.3 & 83.9 & 172.1\\
Bartlett & 13.1 & 32.6 & 43.8 & 84.4 & 174.0\\
Gaussian & \textbf{13.9} & \textbf{33.3} & \textbf{44.5} & \textbf{84.9} & \textbf{176.6}\\
\hline
\hline
\end{tabular}
 }
\end{table}

\subsection{Ablation Study}

\textbf{GMMFormer Block.} For ablations on the proposed GMMFormer block, we first alternate the proposed network into a baseline by replacing GMMFormer blocks with vanilla Transformer blocks and removing the query diverse loss. As illustrated in Table \ref{tab:ablation}, keeping the GMMFormer block for the baseline model will improve retrieval performance, and replacing it will degrade retrieval performance compared to the full setup, demonstrating its effectiveness for PRVR. We owe it that the GMMFormer block can provide multi-scale clip information and perceive video moments with different lengths.

\begin{figure}[t]
  \centering
  \includegraphics[width=0.8\linewidth]{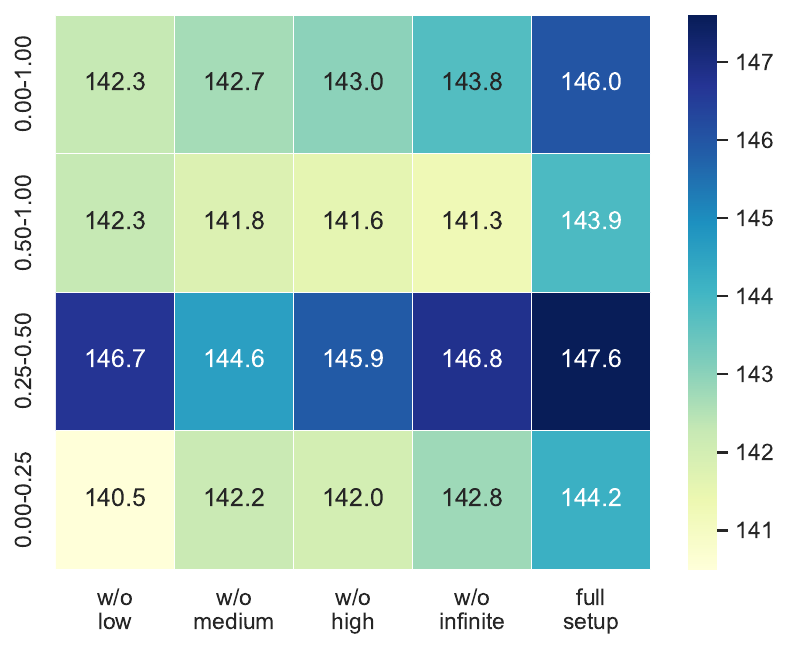}
  \caption{Ablation studies of the Gaussian block on ActivityNet Captions with different types of queries. Queries are grouped according to their moment-to-video ratios (M/V). Different Gaussian blocks are good at handling different M/V groups. And a GMMFormer variant w/o any Gaussian block will perform poorly on the corresponding group.}
  \label{gb}
\end{figure}

\textbf{Gaussian Block.} In Section \ref{sec33}, we choose four types of Gaussian blocks with low, medium, high and infinite variance respectively to perceive different-length video moments. In this subsection, we investigate the impact of these Gaussian blocks. We successively remove one kind of these Gaussian blocks and construct four variants (\ie, w/o low, w/o medium, w/o high and w/o infinite). Then, we define the moment-to-video ratio (M/V) of a query measured by its corresponding moment’s length ratio in the entire video. Next, we split ActivityNet Captions into four groups according to M/V (\ie, 0.00-0.25, 0.25-0.50, 0.50-1.00, 0.00-1.00). We report the performance (SumR) of different variants on different groups in Figure \ref{gb}. All variants perform worse than the full setup, showing that four types of Gaussian blocks all play their roles in GMMFormer. Interestingly, we find that in the group with low M/V (0.00-0.25), the variant w/o low is the worst performer. The same phenomenon happens to the variant w/o medium in the group with medium M/V (0.25-0.50) and the variants w/o high or infinite in the group with high M/V (0.50-1.00), verifying the rationality of designed multi-scale Gaussian blocks.

\textbf{Constraint Window.} We also investigate the design of the constraint window during frame interactions. Specifically, we alternate three types of constraint windows (\ie, Boxcar, Bartlett, Gaussian) and report their performance in Table \ref{tab:window}. As can be seen, the variant with the Boxcar window performs poorly, which is consistent with the intuition that video frames should pay more attention to adjacent frames. Besides, the Gaussian window outperforms the Bartlett window. We attribute this to the smooth and natural characteristics of the Gaussian distribution.

\textbf{Query Diverse Loss.} We provide ablations on the proposed query diverse loss for PRVR in Table \ref{tab:ablation}. Compared to the full setup, removing query diverse loss will degrade retrieval performance and adding it to the baseline will improve retrieval performance, proving its effectiveness for the PRVR task.

\subsection{Qualitative Results}

\begin{figure} [t]
  \centering
  \subfloat[]{\includegraphics[width = 0.16\textwidth]{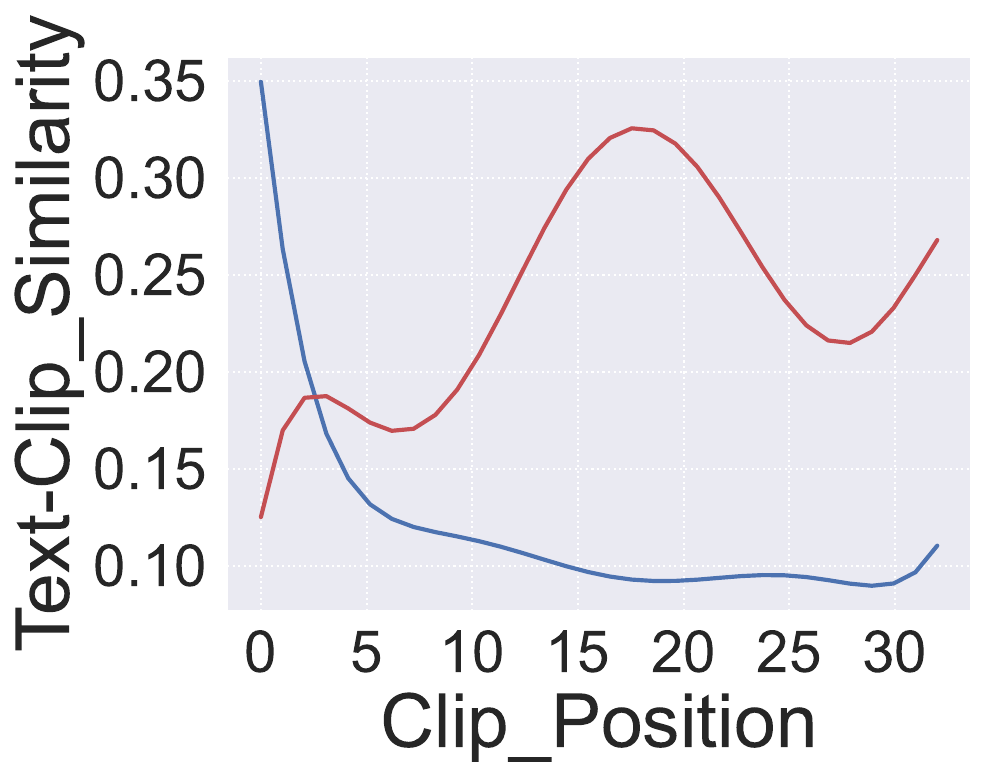}}
  \subfloat[]{\includegraphics[width = 0.16\textwidth]{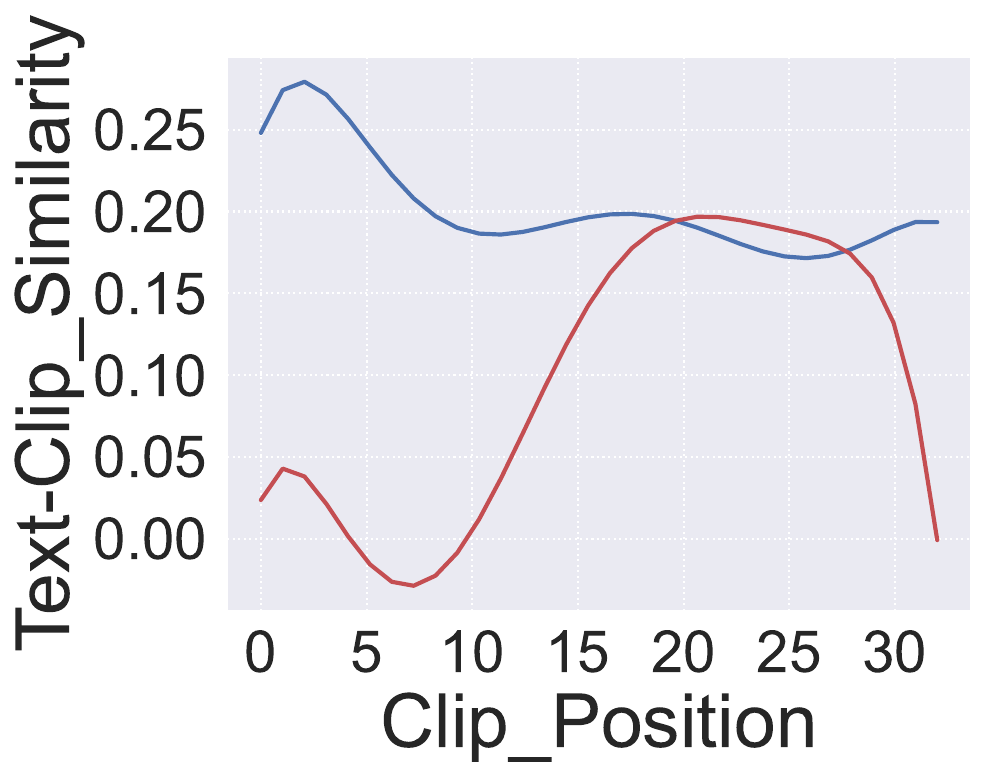}}
  \subfloat[]{\includegraphics[width = 0.16\textwidth]{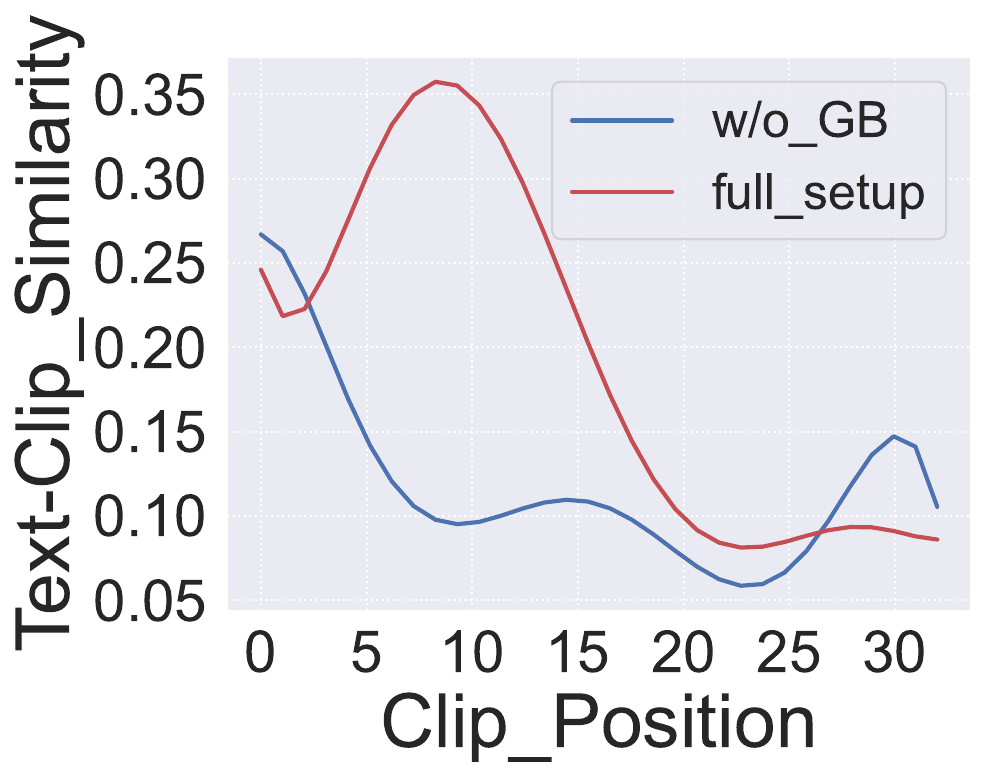}}
  \caption{Text-clip similarity visualizations on TVR. w/o GB means a variant of GMMFormer replacing GMMFormer blocks with vanilla Transformer blocks. Note that we smooth out similarity intervals for better observation.}
  \label{var}
\end{figure}

\textbf{Text-Clip Similarity.} To further reveal the ability of the designed GMMFormer block to explore the partial relevance between videos and texts, we present several text-clip similarity examples on TVR. Specifically, we replace GMMFormer blocks in GMMFormer with vanilla Transformer blocks to build a baseline called w/o GB. As illustrated in Figure \ref{var}, the model with GMMFormer blocks can generate more discriminative clip embeddings. For example, in Figure \ref{var} (a), the model w/o GB fails to localize the moment relevant to the text. And in Figure \ref{var} (b) and (c), the model w/o GB confuses different moments while the model with GMMFormer blocks accurately distinguishes between relevant and irrelevant moments.

\begin{figure}[t]
  \centering
  \subfloat[w/o query diverse loss]{\includegraphics[width = 0.23\textwidth]{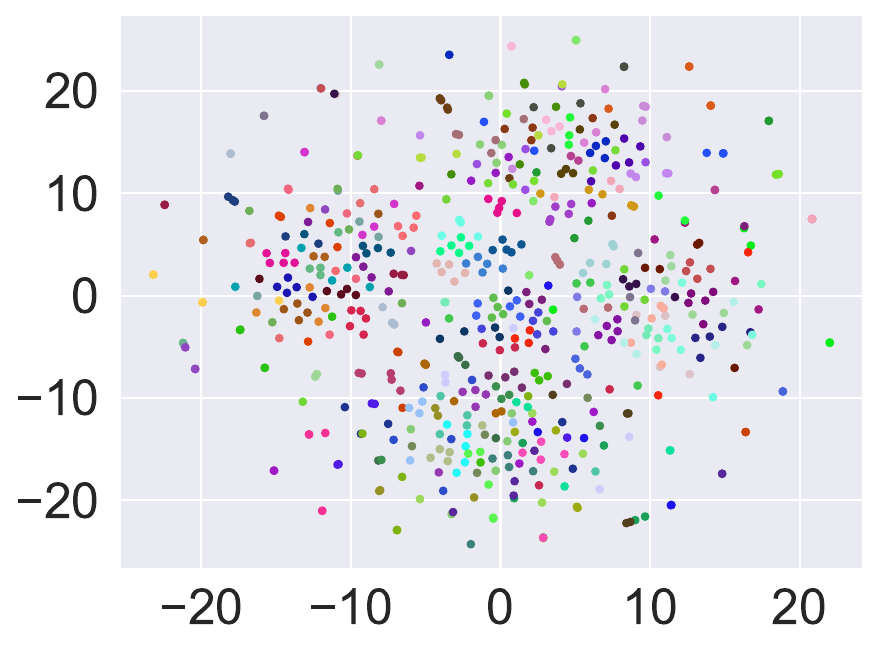}}
  \subfloat[full setup]{\includegraphics[width = 0.23\textwidth]{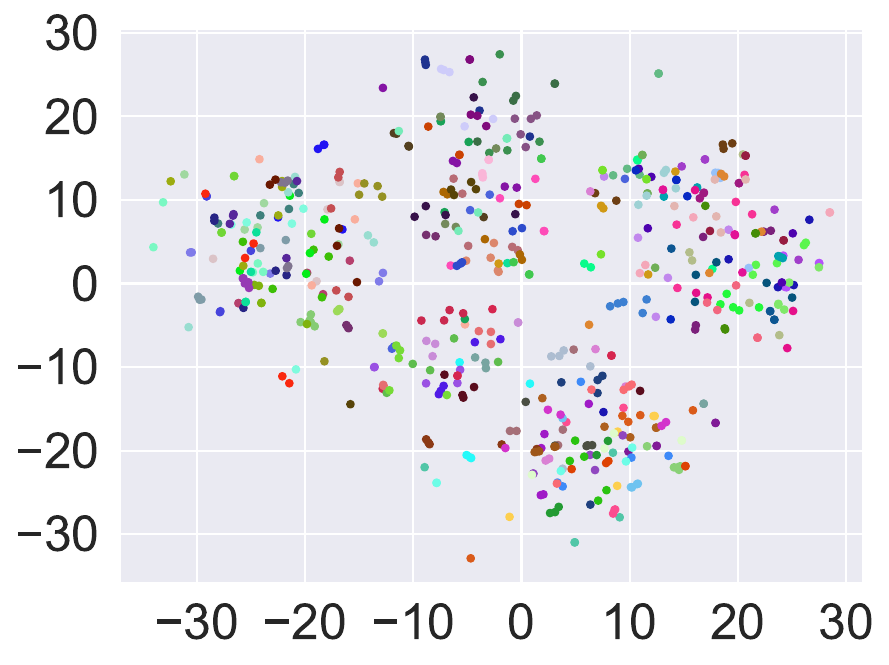}}
  \caption{t-SNE visualizations on the TVR subset. Texts relevant to different / same videos are marked with different / same colors.}
  \label{tsne}
\end{figure}

\textbf{t-SNE Visualization.} To further reveal the ability of the designed query diverse loss to preserve semantic structure of text representations, we show some t-SNE visualizations of GMMFormer without query diverse loss and the full setup. We randomly sample a small subset of videos with their corresponding text queries on TVR for better observation. As shown in Figure \ref{tsne}, the model with the query diverse loss can aggregate relevant text embeddings to a greater extent and make the entire embedding space more discriminative.

\section{Conclusions}

This paper proposes GMMFormer, a Gaussian-Mixture-Model based Transformer for the PRVR task. GMMFormer incorporates a Gaussian-Mixture-Model constraint to model clip representations implicitly and generates compact clip embeddings with high information density. Besides, we propose a query diverse loss to distinguish text queries relevant to the same video, preserving the semantic structure of text representations. Extensive experiments and ablation studies on three large-scale video datasets demonstrate the effectiveness and efficiency of our GMMFormer. In particular, GMMFormer is about 2.5 times faster than the past SOTA MS-SL and the storage overhead of GMMFormer is 20 times smaller than MS-SL.

\setcounter{secnumdepth}{0}

\section{Acknowledgments}

This work is supported in part by the National Natural Science Foundation of China under grant 62171248, 62301189,  Guangdong Basic and Applied Basic Research Foundation under grant 2021A1515110066, Guangdong Provincial Key Laboratory of Novel Security Intelligence Technologies (2022B1212010005), the PCNL KEY project (PCL2023AS6-1), and Shenzhen Science and Technology Program under Grant JCYJ20220818101012025, RCBS20221008093124061, GXWD20220811172936001.

\bibliography{aaai24}


\appendix
\clearpage
\newpage
\onecolumn

\setcounter{table}{7}
\setcounter{figure}{6}

\noindent\textbf{\LARGE{Appendix}}

\section{More Qualitative Results}

\begin{figure*}[t]
  \centering
  \includegraphics[width=\linewidth]{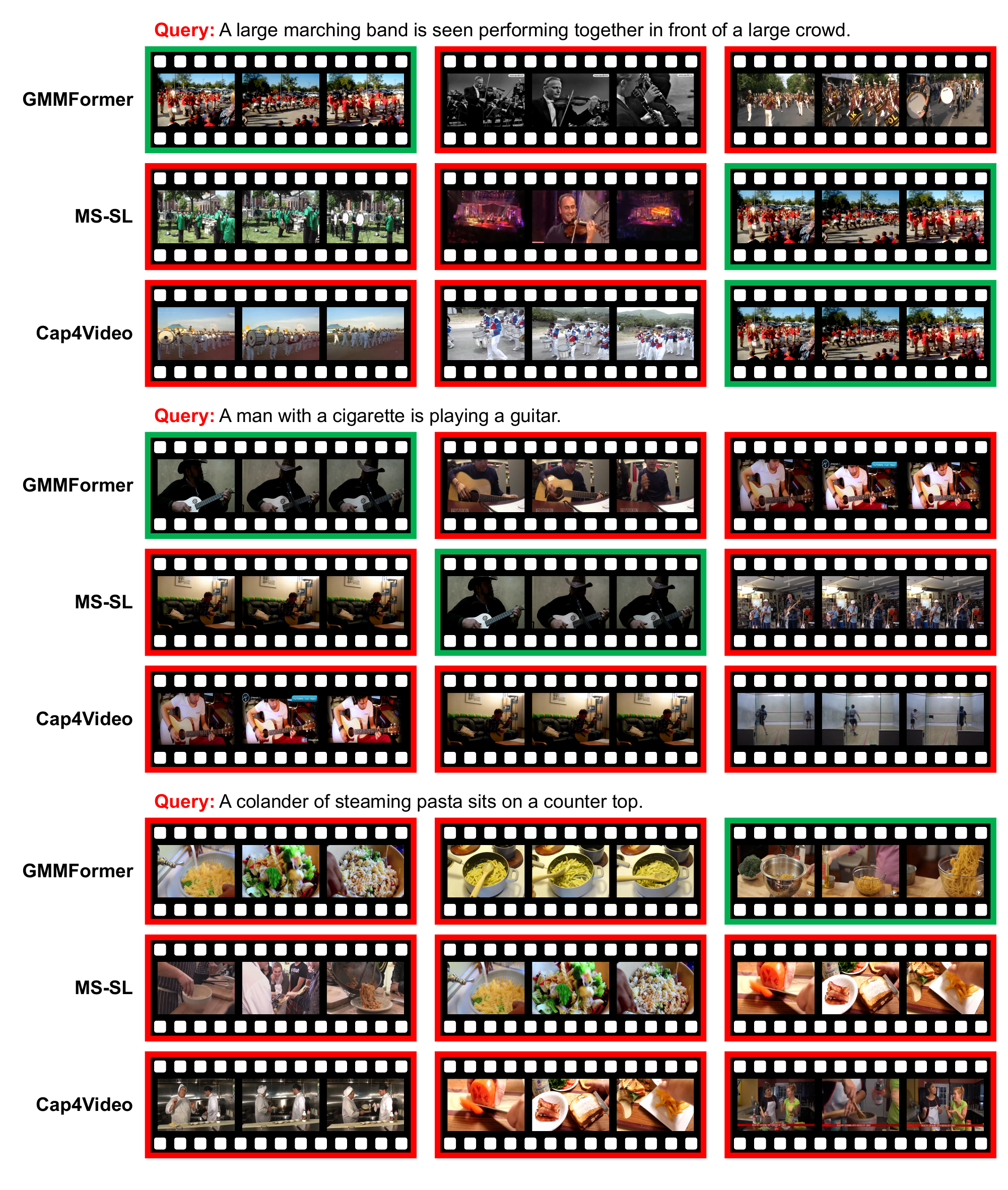}
  \caption{Text-to-video retrieval results on ActivityNet Captions. We show top three retrieval results for each query. We use green boxes to represent the video corresponding to the query and red boxes to represent the irrelevant videos that retrieved.}
  \label{vis}
\end{figure*}

\subsection{Qualitative Retrieval Results}

To qualitatively validate the effectiveness of GMMFormer, we display several typical examples on ActivityNet Captions in Figure \ref{vis}. Based on these retrieval results, we could see that our GMMFormer model can return more precise retrieval results than other competitive models (\ie, MS-SL, Cap4Video).

\section{More Implementation Details}

We set $M_c$ to 32 when downsampling and the maximum frame number $M_f$ to 128. Once the number of frames exceeds $M_f$, it will be uniformly downsampled to $M_f$. For sentences, we set the maximum length of query words $N$ to 30 on TVR and Charades-STA, and 64 on ActivityNet Captions. The words outside the maximum length in a sentence will be discarded. For the Transformer module, we set its hidden size $d=384$, and four attention heads are employed. For model training, we utilize an Adam optimizer with a mini-batch size of 128. The number of epochs is set to 100. Our model is implemented in Pytorch with an Nvidia RTX3080Ti GPU. Other detailed hyper-parameter settings are shown in Table \ref{param}. During training, we take a learning rate adjustment schedule for the learning rate, similar to XML.

\begin{table} [h]
\caption{Hyper-parameter settings of TVR, Activity-Captions and Charades-STA.
}
\label{param}
\centering 
\scalebox{0.88}{
\begin{tabular}{l|c|c|c}
\hline
\textbf{Params} & \textbf{TVR} & \textbf{ActivityNet-Captions} & \textbf{Charades-STA} \\
\hline
learning rate & 3e-4 & 2.5e-4 & 2.5e-4 \\
$\alpha_{v}$ & 0.3 & 0.3 & 0.3 \\
$\alpha_{c}$ & 0.7 & 0.7 & 0.7 \\
$\alpha$ & 32 & 32 & 32 \\
$\delta$ & 0.15 & 0.2 & 0.15 \\
$m$ & 0.1 & 0.2 & 0.2 \\
$\lambda_1$ & 5e-2 & 2e-2 & 2e-2 \\
$\lambda_2$ & 4e-2 & 4e-2 & 2e-2 \\
$\lambda_3$ & 1e-3 & 1.5e-2 & 5e-3 \\
\hline
\hline
\end{tabular}
 }
\end{table}

\end{document}